\documentclass[journal]{IEEEtran}

\usepackage[T1]{fontenc}
\usepackage{amsmath,amssymb}
\usepackage{booktabs}
\usepackage{cite}
\usepackage{graphicx}
\usepackage{url}
\usepackage{microtype}

\usepackage{etoolbox}

\patchcmd{\subsection}
  {3.5ex plus 1.5ex minus 1.5ex}
  {2.9ex plus 1.5ex minus 1.5ex}
  {}
  {\PackageWarning{spacing}{Subsection spacing patch failed}}

\newcommand{\tableformat}{%
  \renewcommand{\arraystretch}{1.12}%
  \setlength{\tabcolsep}{4.2pt}%
}
\newcommand{\best}[1]{\textbf{#1}}
\newcommand{\second}[1]{\underline{#1}}

\newcommand{\authorfigure}[4]{%
  \IfFileExists{figures/#1.pdf}{\includegraphics[width=#2]{figures/#1.pdf}}{%
    \IfFileExists{figures/#1.png}{\includegraphics[width=#2]{figures/#1.png}}{%
      \fbox{\parbox[c][#3][c]{\dimexpr#2-2\fboxsep-2\fboxrule\relax}{\centering\footnotesize #4}}%
    }%
  }%
}

\title{QSCP: Beyond Class-Name Prompts for Query-Guided Semantic Change Parsing}

\author{Yuan Qian, Jie Ma%
\thanks{This work was supported in part by the Natural Science Foundation of China under Grant 62101052; and in part by the Fundamental Research Funds for the Central Universities under Grant 2024JJ040. (Corresponding author: Jie Ma.)}%
\thanks{The authors are with the School of Information Science and Technology, Beijing Foreign Studies University, Beijing 100089, China (e-mail: majie\_sist@bfsu.edu.cn).}%
}

\begin{document}
\raggedbottom
\setlength{\abovedisplayskip}{7pt plus 2pt minus 2pt}
\setlength{\belowdisplayskip}{7pt plus 2pt minus 2pt}
\maketitle

\begin{abstract}
Traditional change detection (CD) identifies changes between bi-temporal remote sensing images, while semantic change detection (SCD) assigns predefined land-cover classes. However, mapping all changes may not meet a user's specific needs. Referring change detection (RCD) enables selective retrieval through category prompts. 
However, existing category-prompted RCD uses the queried category to specify the
destination of a change and returns only a binary mask
of the corresponding regions. Users may instead request a particular transition and paired semantic maps to understand what changed into what. Such requests require explicit source and target reasoning
beyond target-class localization. To address these needs, we propose query-guided semantic change parsing (QSCP), which supports category names, synonyms, and intent-bearing sentences and returns a query-specific mask with paired temporal semantic maps. QSCP parses requests into intents and semantic slots, composes bidirectional visual evidence, and predicts both temporal states with a query-conditioned decoder.  On SECOND, QSCP outperforms RCDNet on synonym, sentence, and transition queries and improves end-to-end semantic prediction over evaluated semantic baselines. WHU-CDC experiments further assess cross-dataset transfer and consistency across equivalent expressions without target-domain training. Code is available at \url{https://github.com/qianyuancs/QSCP}.

\end{abstract}

\begin{IEEEkeywords}
Referring change detection, semantic change detection, natural language queries.
\end{IEEEkeywords}

\section{Introduction}
\IEEEPARstart{C}{hange} detection (CD) supports land-use monitoring, urban development analysis, and disaster response by identifying changes in bi-temporal imagery \cite{lu2004change,hamidi2023flood}. Deep Siamese networks extract spatial and temporal evidence \cite{daudt2018siamese}. Transformer-based methods further model long-range
spatial and temporal context
\cite{chen2022bit,bandara2022changeformer}. Semantic
change detection (SCD)  assigns land-cover labels to changed regions \cite{yang2022second,ding2024scannet}. Yet a complete change map may not answer a specific request, such as locating only the areas where buildings replaced vegetation.

Referring change detection (RCD) introduces category prompts to retrieve changes of interest \cite{korkmaz2026rcd}. Language-guided methods extend change segmentation to descriptive instructions \cite{liu2025lgcd,jia2026changelisa}. These developments make CD more accessible to users with different information needs. In the target-class RCD formulation, the response is a binary mask of changes into the queried category. Users may instead ask where buildings disappeared (\emph{from-class}), where ponds became buildings (\emph{transition}), or where buildings appeared or disappeared (\emph{involved-class}). They may also use ``houses'' instead of ``buildings''. A target category alone cannot distinguish these temporal relations. Moreover, users may want both temporal semantic maps: a mask of new buildings does not reveal whether they replaced water or vegetation.

These requests couple localization and classification. A transition requires source and destination evidence to agree at the same pixels, while semantic prediction must recover states that the query may leave unspecified. DynamicEarth and Seg2Change use text-defined categories for change analysis \cite{li2026dynamicearth,su2026seg2change}; SCD predicts temporal labels without an explicit user-intent interface \cite{ding2024scannet}. RCD can also be adapted to SCD \cite{korkmaz2026rcd}, but categorywise prediction alone does not specify how a relational query selects a region and its two semantic maps. We address this combined output.

We propose \emph{query-guided semantic change parsing} (QSCP). Parsed intents and semantic slots control directional evidence composition; a query-conditioned decoder then predicts both temporal states in the selected region. Language specifies the relation, while vision determines where it occurs and what it connects.
\begin{figure*}[t]
  \centering
  \authorfigure{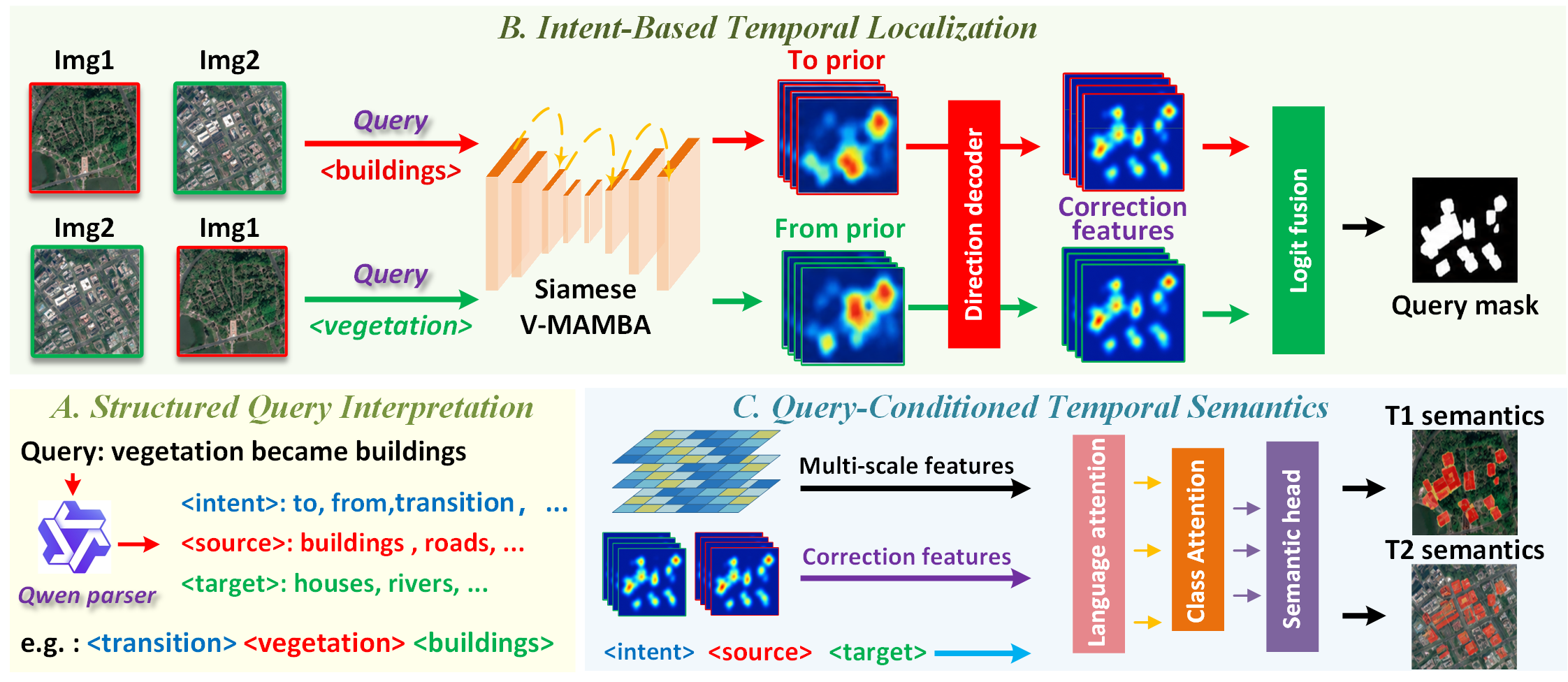}{0.86\textwidth}{84mm}{QSCP framework: author-supplied artwork}
  \caption{QSCP overview. Parsed requests control directional evidence composition and temporal semantic decoding. The predicted mask gates both semantic maps.}
  \label{fig:framework}
\end{figure*}
Our contributions are threefold:
\begin{itemize}
  \setlength{\itemsep}{0pt}
  \item We introduce a structured query formulation that connects synonyms and temporal expressions to explicit operations on directional change evidence.
  \item We develop query-conditioned temporal semantic decoding with directional supervision, providing paired semantic maps alongside the requested change mask.
  \item We evaluate six query suites, parsing accuracy, and semantic ablations on SECOND, followed by cross-dataset transfer with actual predicted routes on WHU-CDC.
\end{itemize}

\section{Method}
Given bi-temporal images $I_1,I_2\in\mathbb{R}^{H\times W\times3}$ and query $u$, QSCP predicts a change mask $\hat M_u$ and semantic maps $\hat S_{1,u},\hat S_{2,u}$ (Fig.~\ref{fig:framework}). The parser identifies classes and their temporal relation; localization composes directional evidence; semantic decoding recovers both land-cover states.

\subsection{Structured Query Interpretation}
The language stage separates class identity from temporal intent. Given query $u$, a text-only large language model (LLM) receives the ontology $\mathcal C$ and intent definitions $\mathcal Z$, then returns JSON normalized as
\begin{equation}
\begin{aligned}
J_u &= \mathcal P_{\mathrm{LLM}}(u;\mathcal C,\mathcal Z),\\
\pi(u)&=\mathcal N(J_u)=(z,c,s,t,r),
\end{aligned}
\end{equation}
where $\mathcal P_{\mathrm{LLM}}$ denotes text parsing and
$\mathcal N$ performs alias mapping and schema normalization.
Here, $z$ is the intent, $c$ the queried class,
$(s,t)$ the ordered source--target pair, and $r$ the
support flag.

Class intents use $c$; transitions use $(s,t)$. Bare class names default to involved-class. For example, ``ponds became buildings'' specifies water-to-building evidence, whereas ``new buildings'' requires only building-appearance evidence. Unsupported requests ($r=0$) return empty outputs; supported requests require localization. Parsing uses no images or ground-truth correction.

\subsection{Intent-Based Temporal Localization}
Localization applies the same pretrained RCDNet \cite{korkmaz2026rcd} in both image orders:
\begin{equation}
\begin{aligned}
R_c^{\mathrm{to}} &= \mathcal R(I_1,I_2;c),\\
R_c^{\mathrm{from}} &= \mathcal R(I_2,I_1;c),
\end{aligned}
\label{eq:bidirectional_evidence}
\end{equation}
where $\mathcal R$ denotes RCDNet conditioned on the canonical
prompt for class $c$.
The maps encode changes into and out of $c$ in the same spatial coordinates. Frozen CLIP \cite{radford2021clip} encodes the class prompt as a 512-D vector.

A direction-aware decoder jointly refines these estimates
using bi-temporal features.
At each scale, a $1\times1$ Conv--BN--ReLU block projects
concatenated temporal features and their absolute difference
to 96 channels.
After bilinear alignment to the finest scale, two additional
branches encode the absolute RGB difference and the
directional prior:
\begin{equation}
\begin{aligned}
\Delta I&=|I_2-I_1|,\\
B_c&=\operatorname{Concat}
\bigl(R_c^{\mathrm{to}},R_c^{\mathrm{from}},
|R_c^{\mathrm{to}}-R_c^{\mathrm{from}}|\bigr).
\end{aligned}
\label{eq:direction_inputs}
\end{equation}
The RGB and prior branches use two and one
$3\times3$ Conv--BN--ReLU blocks, respectively.
A 96-D linear projection of $e_c$ is broadcast and added to each branch.
All branches are concatenated and fused by two
$3\times3$ Conv--BN--ReLU blocks.
A $1\times1$ classifier produces three logits, which are
upsampled to the input resolution and normalized:
\begin{equation}
\mathbf D_c
=\operatorname{softmax}(\mathbf L_c)
=\bigl(D_c^{\mathrm{other}},
D_c^{\mathrm{to}},D_c^{\mathrm{from}}\bigr).
\label{eq:direction_probabilities}
\end{equation}

To retain the pretrained evidence, we fuse each directional
probability with its RCDNet counterpart in binary log-odds space:
\begin{equation}
P_c^d=\sigma\!\left(
(1-\alpha)\ell(R_c^d)+\alpha\ell(D_c^d)
\right),
\qquad d\in\{\mathrm{to},\mathrm{from}\}.
\label{eq:fusion}
\end{equation}
Here, $\sigma$ is sigmoid and $\ell(p)=\log[p/(1-p)]$, with probabilities clipped away from zero and one. Binary log-odds place both estimates on a common scale; validation selects $\alpha$.

The parsed intent then determines the query score:
\begin{equation}
A_u=
\begin{cases}
0, & r=0,\\
P_c^{\mathrm{to}}, & r=1,\ z=\mathrm{to},\\
P_c^{\mathrm{from}}, & r=1,\ z=\mathrm{from},\\
P_c^{\mathrm{to}}+P_c^{\mathrm{from}},
& r=1,\ z=\mathrm{involved},\\
P_t^{\mathrm{to}}P_s^{\mathrm{from}},
& r=1,\ z=\mathrm{transition}.
\end{cases}
\label{eq:router}
\end{equation}
The sum retains either direction but is not a normalized probability, not a normalized probability; multiplication requires spatial agreement. Thresholding and small-component removal yield $\hat M_u$, with score-specific calibration (Section~\ref{sec:implementation}).

\begin{table*}[t]
\centering
\caption{SECOND query comparison (\%): suite BIoU and Present-class temporal semantic mIoU. Sem.GT uses GT regions; E2E-Sem includes predicted-mask errors. Bold: best; N/A: unavailable.}
\label{tab:query_comparison}
\footnotesize
\tableformat
\begin{tabular*}{\textwidth}{@{\extracolsep{\fill}}lcccccccc@{}}
\toprule
Method
& All-class
& Present-class
& Synonym
& Sentence
& Transition
& Unknown
& Sem.GT
& E2E-Sem \\
\midrule
RCDNet
& 71.15
& 70.56
& 53.34
& 53.56
& 48.56
& 49.65
& N/A & N/A \\
\addlinespace[2pt]
SCanNet
& 64.32
& 64.43
& 44.87
& 46.09
& 48.16
& \best{100.00}
& 33.36 & 22.19 \\
\addlinespace[2pt]
Seg2Change
& 66.64
& 65.27
& 45.10
& 46.29
& 48.16
& \best{100.00}
& 44.71 & 23.12 \\
\addlinespace[2pt]
DynamicEarth
& 59.70
& 58.16
& 45.22
& 46.43
& 48.16
& \best{100.00}
& N/A & N/A \\
\midrule
QSCP (ours)
& \best{71.35}
& \best{70.71}
& \best{70.71}
& \best{71.74}
& \best{70.99}
& \best{100.00}
& \best{62.35}
& \best{38.01} \\
\bottomrule
\end{tabular*}
\par\vspace{3pt}\noindent
\parbox{\textwidth}{\footnotesize
RCDNet encodes raw text; SCanNet, Seg2Change, and DynamicEarth use the canonical-keyword adapter. QSCP includes parsing.
Unknown BIoU is not query-level rejection accuracy.}
\end{table*}

\subsection{Query-Conditioned Temporal Semantics}
The semantic decoder predicts temporal classes not fully
specified by the query. It fuses multi-scale temporal features
with correction features and directional logits:
\begin{equation}
X=\Phi([\{G_l\}_l,\mathcal Q_p(C_p),
\mathcal Q_s(C_s),\mathcal Q_d([L_p,L_s])]),
\end{equation}
where $G_l$ projects $[F_1^l,F_2^l,|F_2^l-F_1^l|]$ through
a $1\times1$ Conv--BN--ReLU block.
The $Q$ branches project 96-channel correction features
and concatenated directional logits through $3\times3$
Conv--BN--ReLU blocks to 128 channels on a $64\times64$
grid. $\Phi$ concatenates these with $\{G_l\}$ and applies
two $3\times3$ blocks. For transitions, $p/s$ index
target/source branches; otherwise, $p$ uses the queried
class and source inputs are zero.

Five 128-D tokens $T_u$ encode query text, class, source,
target, and intent using projected CLIP vectors and a learned
intent embedding. Unused class vectors are zeroed before
projection. Two four-head cross-attention blocks\cite{vaswani2017attention} yield
\begin{equation}
\begin{aligned}
H_0&=\operatorname{Flatten}(X)+\operatorname{Mean}(T_u),\\
H_1&=\operatorname{LN}(H_0+\operatorname{MHA}(H_0,T_u,T_u)),\\
H_2&=\operatorname{LN}(H_1+\operatorname{MHA}(H_1,E,E)),
\end{aligned}
\end{equation}
where the mean is broadcast spatially, $E$ contains projected
class embeddings, and MHA arguments are query, key, and value.
The two blocks have separate parameters.

Each temporal head uses $3\times3$ Conv--BN--ReLU,
dropout (0.1), and a $1\times1$ seven-class classifier.
Upsampled logits yield argmax labels within $\hat M_u$
and background elsewhere, leaving the mask unchanged.

\subsection{Semantic Supervision}
We derive each query mask from valid changed pixels in the
temporal annotations. To-class and from-class queries select
pixels by their target and source labels, respectively;
involved-class queries take their union; transitions
require both labels to match. Within this mask, the semantic
decoder is supervised using both temporal labels.
Gold query slots define supervision and evaluation masks;
predicted slots alone drive inference.

However, a transition mask contains only the specified
source--target pair, providing limited semantic diversity.
We therefore add directional supervision: for $s\to t$,
the first-time prediction is supervised on all changes
into $t$, and the second-time prediction on all changes
out of $s$. These broader regions expose the decoder
to alternative origins and destinations.
The objective is
\begin{equation}
\mathcal L=\mathcal L_{\rm query}
+1.2\mathcal L_{\rm direction}.
\end{equation}
Each term sums the losses at both times, combining
class-weighted cross entropy\cite{lin2017focal}, focal loss
($\gamma=1.5$), and Dice loss with coefficients
$1$, $0.4$, and $0.3$, respectively.
Pixels outside the supervision region are ignored.
RCDNet, CLIP, and the direction decoder remain frozen.

\begin{table}[t]
\caption{SECOND whole-image CD (\%): one merged map per image. Bold/underline: best/second best.}
\label{tab:whole}
\centering
\footnotesize
\tableformat
\begin{tabular*}{\columnwidth}{@{\extracolsep{\fill}}lccc@{}}
\toprule
Method & Changed IoU & Unchanged IoU & BIoU \\
\midrule
SCanNet \cite{ding2024scannet} & \best{57.47} & \second{87.65} & \best{72.56} \\
Seg2Change \cite{su2026seg2change} & 48.61 & 86.84 & 67.72 \\
DynamicEarth \cite{li2026dynamicearth} & 38.36 & 80.83 & 59.60 \\
Native RCDNet \cite{korkmaz2026rcd} & 57.17 & 87.60 & 72.39 \\
\midrule
QSCP & \second{57.25} & \best{87.78} & \second{72.51} \\
\bottomrule
\end{tabular*}
\end{table}

\section{Experiments}
\subsection{Datasets}
\textit{SECOND} contains 4,662 bi-temporal image pairs of $512\times512$ pixels, which we split into 2,672 training,
296 validation, and 1,694 test pairs, preserving the
original test set \cite{yang2022second}.
Its six foreground classes are low vegetation,
non-vegetated ground surface, tree, water, building,
and playground.

\textit{WHU-CDC} provides building-change imagery \cite{ji2019whu}. We test 744 pairs (181 changed, 563 unchanged) with SECOND weights and calibration fixed, without target-domain training or model selection.

\subsection{Query Bank Construction}
As in template-based instruction construction~\cite{lai2024lisa}, class names, aliases, and intent templates are paired with annotation-derived masks.

\textit{SECOND.}
Training uses 120 expressions and 34,595 image--query pairs. All-class tests all six names per image (10,164 queries, including 3,870 empty targets); Present-class retains the 6,294 nonempty targets. Synonym, Sentence, and Transition contain 18,882, 32,512, and 17,632 positive queries; Unknown has 1,694 empty targets.

\textit{WHU-CDC.}
A fixed bank of 19 expressions is paired with 744 images:
one canonical building query, four synonyms, ten sentences,
and four unsupported queries.
The 15 supported expressions refer to the same
involved-building request and share the binary
building-change GT; unsupported queries test ontology
rejection. Binary labels do not support directional
transition evaluation.
QSCP uses predicted routes with the SECOND parser
configuration, while baselines receive the same raw
expressions.

\begin{table}[t]
\caption{WHU-CDC transfer (\%). Baselines encode raw text; QSCP uses normalized routes. Mean: supported-suite BIoU; Unknown: rejection accuracy.}
\label{tab:whu}
\centering
\footnotesize
\tableformat
\setlength{\tabcolsep}{3pt}
\begin{tabular*}{\columnwidth}{@{\extracolsep{\fill}}lccccc@{}}
\toprule
Method & Canonical & Synonym & Sentence & Mean & Unknown \\
\midrule
RCDNet & \second{71.06} & 62.39 & 50.07 & 61.17 & 54.77 \\
DynamicEarth & 60.74 & 57.96 & \second{57.39} & 58.70 & 27.92 \\
Seg2Change & \best{78.11} & \best{73.32} & 54.88 & \second{68.77} & \second{98.92} \\
\midrule
QSCP & 70.94 & \second{70.94} & \best{70.94} & \best{70.94} & \best{100.00} \\
\bottomrule
\end{tabular*}
\end{table}

\begin{figure*}[t]
  \centering
  \authorfigure{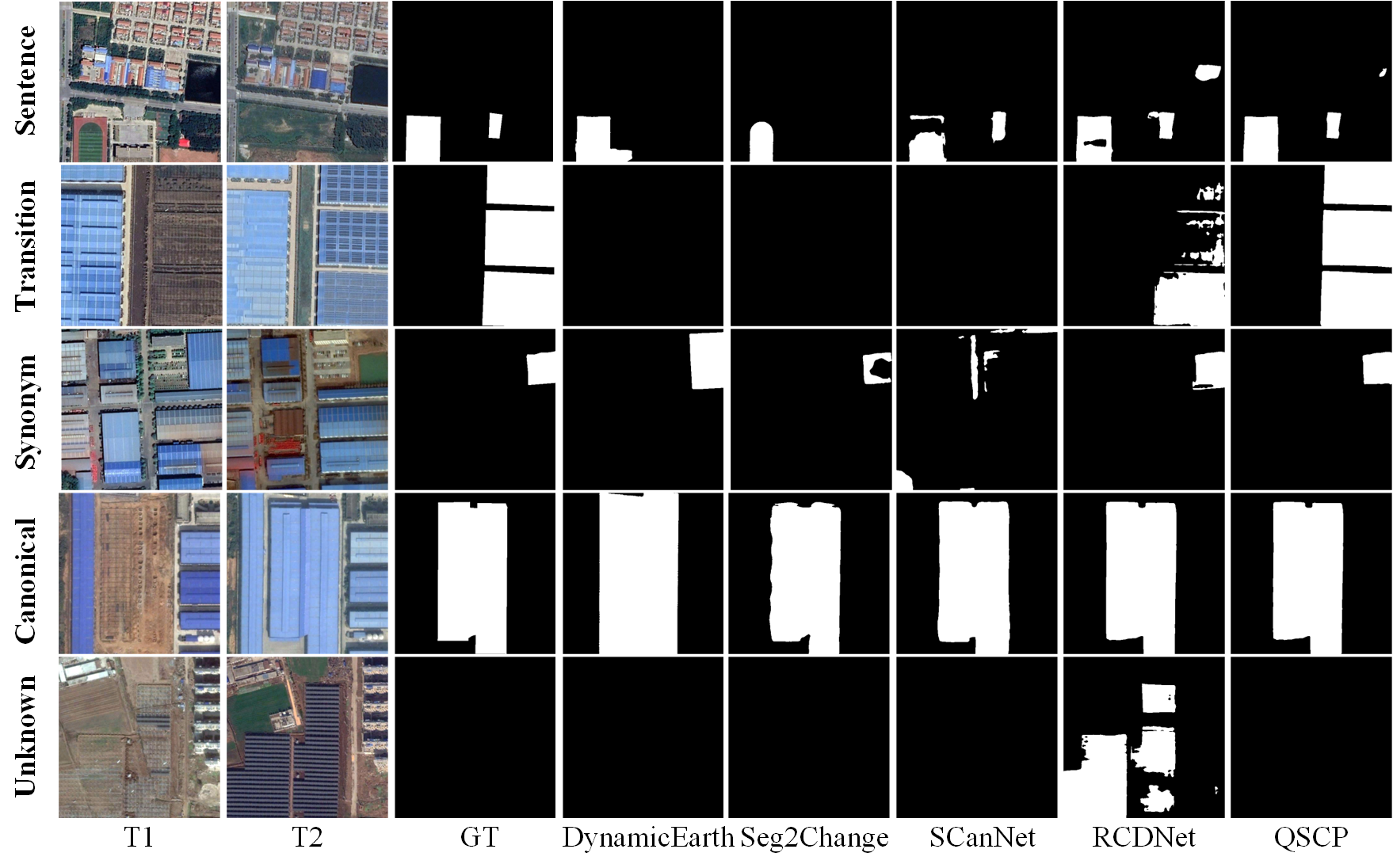}{0.82\textwidth}{100mm}{Change-mask comparison: author-supplied artwork}
\caption{Qualitative comparisons on SECOND.
From top to bottom, the queries are
``playground-related changes'' (Sentence),
``paved ground became houses'' (Transition),
``water surface'' (Synonym),
``building'' (Canonical), and
``oil tanks'' (Unknown).
White indicates the queried change regions.
The unknown query tests ontology rejection.
Examples are selected for illustration.}
  \label{fig:change-comparison}
\end{figure*}

\subsection{Implementation and Calibration}
\label{sec:implementation}
Experiments use one NVIDIA RTX 4090 GPU with 24~GB memory, the official SECOND RCDNet checkpoint, CLIP ViT-B/32, and native SECOND normalization. Decoders are trained in stages; inference removes components smaller than 64 pixels at $512\times512$ resolution.

Parsing uses \texttt{qwen-plus} at temperature 0 with fixed alias/schema normalization and cached responses for repeated texts. WHU retains the SECOND prompt and rules, with five texts per API request.

Fusion and decision thresholds are calibrated on the
SECOND validation split and fixed before testing.
The selected fusion weight is $\alpha=0.25$.
The whole-image, class-intent, and transition thresholds
are $0.60$, $0.55$, and $0.20$, respectively.
Separate thresholds accommodate the different score
constructions, particularly the product used for transitions.
The same settings are used on WHU-CDC without recalibration. For semantic refinement, cross-entropy and focal losses
use fixed class weights $(0,1.15,1,1.1,2.4,1,2.2)$,
ordered as background followed by the six foreground classes.

\subsection{Evaluation}
BIoU averages pooled changed and unchanged IoU: one merged map per image in Table~\ref{tab:whole}, or all image--query pairs in Table~\ref{tab:query_comparison}. Sem.GT and E2E-Sem average temporal foreground-class mIoUs; only E2E-Sem includes missed and spurious regions. Paired bootstrap resamples images with their queries, conditioning on fixed models and banks.

\subsection{Whole-Image Change Detection}

QSCP retains competitive CD performance (Table~\ref{tab:whole}): second-best BIoU (72.51), highest unchanged IoU, and a 0.13-point gain over RCDNet (95\% CI: $[0.06,0.20]$). Its 0.04-point gap to SCanNet has an interval spanning zero. Adding query and semantic outputs therefore preserves whole-image localization.

\subsection{Query-Based Comparison}
Table~\ref{tab:query_comparison} compares complete QSCP with raw-text RCDNet and keyword-adapted baselines. The latter return empty masks for unmatched expressions; their low language-suite scores measure interface limitations, not native text encoding. Native raw-text baselines are tested on WHU.

On Synonym, Sentence, and Transition, QSCP scores
70.71, 71.74, and 70.99 BIoU versus RCDNet's
53.34, 53.56, and 48.56, respectively.
The mean gain is 19.32 points (95\% CI: $[18.76,19.88]$),
including parsing and routing gains beyond visual refinement.
Unknown measures ontology rejection, not unseen-class recognition.
Fig.~\ref{fig:change-comparison} shows localization examples.

\subsection{Parsing Accuracy}
We evaluate distinct query texts using cached parser outputs
after fixed text normalization.
Exact-route accuracy requires correct intent, support flag,
and active semantic slots.
On SECOND, intent and support accuracy are 100\% across
all suites. Exact-route accuracy is 99.94\% for the
1,549 transition expressions, with one slot error,
and 100\% for the remaining suites.
All 19 WHU-CDC expressions are parsed correctly after
normalization.
These results demonstrate reliable parsing within the
controlled query banks, not unrestricted language understanding.

\subsection{Temporal Semantic Parsing}
\begin{figure}[t]
  \centering
  \authorfigure{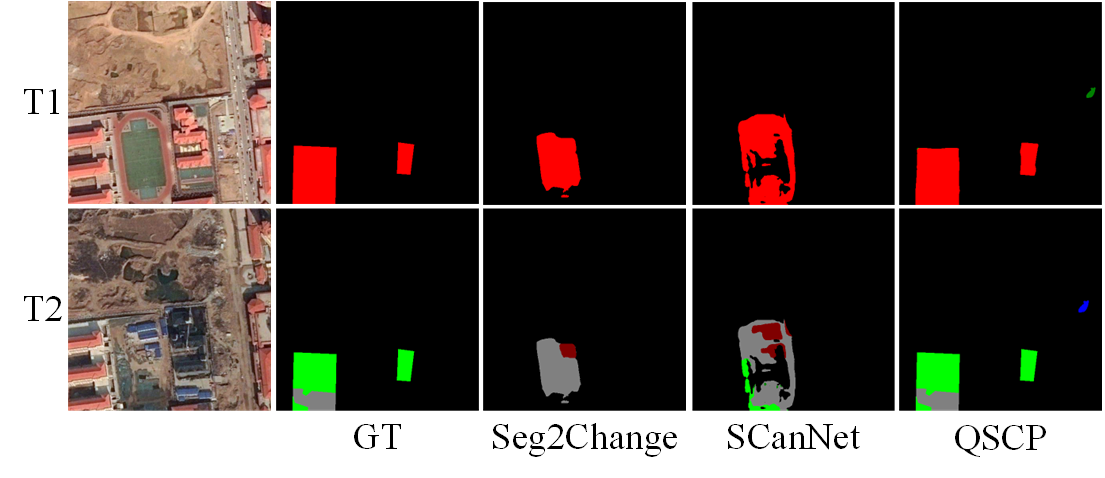}{\columnwidth}{38mm}{Temporal semantic comparison: author-supplied artwork}
  \caption{SECOND temporal semantics at T1/T2. Colors denote classes; black denotes background or query-excluded regions.}
  \label{fig:semantic-comparison}
\end{figure}

QSCP reaches 62.35 Sem.GT and 38.01 E2E-Sem on Present-class (Table~\ref{tab:query_comparison}), exceeding Seg2Change's 44.71/23.12 and SCanNet's 33.36/22.19. Lower E2E scores expose localization errors omitted by Sem.GT. Transition Sem.GT is 84.54, but both slots are supplied and positive-mask IoU is 28.03; this is conditioned interpretation, not unconstrained recognition. Fig.~\ref{fig:semantic-comparison} shows both temporal predictions.

\subsection{Cross-Dataset Transfer}

In Table~\ref{tab:whu}, all 15 supported expressions normalize to the same \emph{involved-building} route, giving identical masks and GT per image. The 1/4/10 expressions per suite therefore multiply TP/FP/FN/TN by 1/4/10, as verified in saved counts. IoU ratios are unchanged, explaining the three 70.94 scores. This is consistency across equivalent requests, not independent localization gains.

QSCP achieves the highest sentence-query BIoU, surpassing
RCDNet, Seg2Change, and DynamicEarth by 20.87, 16.06,
and 13.55 percentage points, respectively.
All paired 95\% confidence intervals exclude zero,
supporting its advantage in handling varied expressions
under cross-dataset transfer.
Seg2Change remains stronger on canonical queries and
rejection of supported but absent building changes (58.08\% versus 16.70\%).
QSCP's mean advantage over Seg2Change is 2.16 points,
although its confidence interval spans zero
($[-0.88,5.19]$).

\subsection{Ablation Study}
\textit{Semantic conditioning.} With a frozen localizer (Table~\ref{tab:ablation}), removing explicit language or directional features reduces Sem.GT by 4.87 or 11.10 points. Both aid semantic decoding. The language ablation retains upstream parsing and localization; unchanged mask scores verify their separation from semantics.

\textit{Localization fusion.} We compare the direction decoder alone with its fusion
with pretrained RCD evidence. Fusion raises Present-class/Transition BIoU from 70.14/70.35 to 70.71/70.99 without changing Sem.GT. This modest visual gain is distinct from the full-system language gains.

\begin{table}[t]
\caption{Semantic-head ablation on SECOND validation (\%). Int.: prediction--GT intersection; Pair: temporal-pair accuracy. The localizer is frozen.}
\label{tab:ablation}
\centering
\footnotesize
\tableformat
\begin{tabular}{lcccc}
\toprule
Variant & Sem. GT & Pair GT & Sem. Int. & Mask Pos. \\
\midrule
Full semantic head & \best{71.04} & \best{77.66} & \best{76.92} & 47.95 \\
w/o explicit language & 66.17 & 73.20 & 73.42 & 47.95 \\
w/o direction features & 59.94 & 65.33 & 67.46 & 47.95 \\
\bottomrule
\end{tabular}
\end{table}

\section{Conclusion}
We proposed QSCP to support diverse change queries,
including to-class, from-class, involved-class, and
source-to-target transitions, expressed through category
names, synonyms, or sentences. For each query, QSCP
returns a binary change mask and two temporal semantic
maps, identifying both where the requested change occurs
and the corresponding land-cover states.
Experiments on SECOND demonstrate improved robustness
to controlled language variations and end-to-end semantic
prediction, while preserving competitive change localization.
WHU-CDC results further support cross-dataset transfer
across equivalent query expressions.

\bibliographystyle{IEEEtran}
\bibliography{references}

@article{lu2004change,
  author  = {Dengsheng Lu and Paul Mausel and Eduardo Brond{\'i}zio and Emilio Moran},
  title   = {Change Detection Techniques},
  journal = {Int. J. Remote Sens.},
  volume  = {25},
  number  = {12},
  year    = {2004},
  doi     = {10.1080/0143116031000139863}
}

@article{hamidi2023flood,
  author  = {Ebrahim Hamidi and Brad G. Peter and David F. Mu{\~n}oz and Hamed Moftakhari and Hamid Moradkhani},
  title   = {Fast Flood Extent Monitoring With {SAR} Change Detection Using {Google Earth Engine}},
  journal = {IEEE Trans. Geosci. Remote Sens.},
  volume  = {61},
  pages   = {1--19},
  year    = {2023},
  doi     = {10.1109/TGRS.2023.3240097}
}

@inproceedings{daudt2018siamese,
  author    = {Rodrigo Caye Daudt and Bertrand Le Saux and Alexandre Boulch},
  title     = {Fully Convolutional Siamese Networks for Change Detection},
  booktitle = {Proc. IEEE Int. Conf. Image Process.},
  pages     = {4063--4067},
  year      = {2018}
}

@article{liu2025lgcd,
  author  = {Yixiao Liu and Yizhou Yang and Jinwen Li and Jun Tao and Ruoyu Li and Xiangkun Wang and Min Zhu and Junlong Cheng},
  title   = {{LG-CD}: Enhancing Language-Guided Change Detection Through {SAM2} Adaptation},
  journal = {arXiv preprint arXiv:2509.21894},
  year    = {2025},
  url     = {https://arxiv.org/abs/2509.21894v1}
}

@article{jia2026changelisa,
  author  = {Xiangyu Jia and Zhibo Chen and Shengyi Zhang and Xiaojing Xue},
  title   = {{Change-LISA}: Language-Guided Reasoning for Remote Sensing Change Detection},
  journal = {IEEE Trans. Geosci. Remote Sens.},
  volume  = {64},
  pages   = {1--15},
  year    = {2026},
  doi     = {10.1109/TGRS.2026.3684817}
}

@inproceedings{korkmaz2026rcd,
  author    = {Yilmaz Korkmaz and Jay N. Paranjape and Celso M. de Melo and Vishal M. Patel},
  title     = {Referring Change Detection in Remote Sensing Imagery},
  booktitle = {Proc. IEEE/CVF Winter Conf. Appl. Comput. Vis.},
  pages     = {106--116},
  month     = mar,
  year      = {2026}
}

@article{ji2019whu,
  author = {Shunping Ji and Shiqing Wei and Meng Lu},
  title = {Fully Convolutional Networks for Multisource Building Extraction From an Open Aerial and Satellite Imagery Data Set},
  journal = {IEEE Trans. Geosci. Remote Sens.},
  volume = {57},
  number = {1},
  pages = {574--586},
  year = {2019}
}

@article{li2026dynamicearth,
  author  = {Kaiyu Li and Xiangyong Cao and Yupeng Deng and Chao Pang and Zepeng Xin and Hui Qiao and Tieliang Gong and Deyu Meng and Zhi Wang},
  title   = {{DynamicEarth}: How Far Are We from Open-Vocabulary Change Detection?},
  journal = {Proc. AAAI Conf. Artif. Intell.},
  volume  = {40},
  number  = {8},
  pages   = {6279--6287},
  year    = {2026},
  doi     = {10.1609/aaai.v40i8.37554}
}

@article{su2026seg2change,
  author  = {You Su and Yonghong Song and Jingqi Chen and Zehan Wen},
  title   = {{Seg2Change}: Adapting Open-Vocabulary Semantic Segmentation Model for Remote Sensing Change Detection},
  journal = {arXiv preprint arXiv:2604.11231},
  year    = {2026}
}

@article{yang2022second,
  author  = {Kunping Yang and Gui-Song Xia and Zicheng Liu and Bo Du and Wen Yang and Marcello Pelillo and Liangpei Zhang},
  title   = {Asymmetric Siamese Networks for Semantic Change Detection in Aerial Images},
  journal = {IEEE Trans. Geosci. Remote Sens.},
  volume  = {60},
  pages   = {1--18},
  articleno = {5609818},
  year    = {2022},
  doi     = {10.1109/TGRS.2021.3113912}
}

@article{ding2024scannet,
  author  = {Lei Ding and Jing Zhang and Kai Zhang and Haitao Guo and Bing Liu and Lorenzo Bruzzone},
  title   = {Joint Spatio-Temporal Modeling for Semantic Change Detection in Remote Sensing Images},
  journal = {IEEE Trans. Geosci. Remote Sens.},
  volume  = {62},
  pages   = {1--14},
  year    = {2024},
  doi     = {10.1109/TGRS.2024.3362795}
}

@inproceedings{radford2021clip,
  author    = {Alec Radford and Jong Wook Kim and Chris Hallacy and Aditya Ramesh and Gabriel Goh and Sandhini Agarwal and Girish Sastry and Amanda Askell and Pamela Mishkin and Jack Clark and Gretchen Krueger and Ilya Sutskever},
  title     = {Learning Transferable Visual Models From Natural Language Supervision},
  booktitle = {Proc. Int. Conf. Mach. Learn.},
  volume    = {139},
  pages     = {8748--8763},
  year      = {2021}
}

@inproceedings{lai2024lisa,
  author    = {Xin Lai and Zhuotao Tian and Yukang Chen and Yanwei Li and Yuhui Yuan and Shu Liu and Jiaya Jia},
  title     = {{LISA}: Reasoning Segmentation via Large Language Model},
  booktitle = {Proc. IEEE/CVF Conf. Comput. Vis. Pattern Recognit.},
  pages     = {9579--9589},
  year      = {2024}
}

@article{chen2022bit,
  author  = {Chen, Hao and Qi, Zipeng and Shi, Zhenwei},
  title   = {Remote Sensing Image Change Detection With Transformers},
  journal = {IEEE Transactions on Geoscience and Remote Sensing},
  volume  = {60},
  pages   = {1--14},
  year    = {2022},
  doi     = {10.1109/TGRS.2021.3095166}
}

@inproceedings{bandara2022changeformer,
  author    = {Bandara, Wele Gedara Chaminda and Patel, Vishal M.},
  title     = {A Transformer-Based Siamese Network for Change Detection},
  booktitle = {Proc. IEEE Int. Geosci. Remote Sens. Symp.},
  pages     = {207--210},
  year      = {2022},
  doi       = {10.1109/IGARSS46834.2022.9883686}
}

@inproceedings{lin2017focal,
  author    = {Lin, Tsung-Yi and Goyal, Priya and Girshick, Ross
               and He, Kaiming and Doll{\'a}r, Piotr},
  title     = {Focal Loss for Dense Object Detection},
  booktitle = {Proc. IEEE Int. Conf. Comput. Vis.},
  pages     = {2980--2988},
  year      = {2017},
  doi       = {10.1109/ICCV.2017.324}
}

@inproceedings{vaswani2017attention,
  author    = {Vaswani, Ashish and Shazeer, Noam and
               Parmar, Niki and Uszkoreit, Jakob and
               Jones, Llion and Gomez, Aidan N. and
               Kaiser, {\L}ukasz and Polosukhin, Illia},
  title     = {Attention Is All You Need},
  booktitle = {Advances in Neural Information Processing Systems},
  volume    = {30},
  pages     = {5998--6008},
  year      = {2017}
}

\end{document}